%% file: main.tex
\documentclass[10pt,twocolumn,letterpaper]{article} 

\usepackage{avss}
\usepackage{times}
\usepackage{epsfig}
\usepackage{graphicx}
\usepackage{amsmath}
\usepackage{amssymb}
\usepackage{url}
\usepackage{xcolor}
\usepackage{amsfonts} 
\usepackage{cases}
\usepackage{overpic}
\usepackage{caption}
\usepackage{subcaption}
\usepackage{multicol, multirow}
\usepackage{float}
\usepackage{tabularx}
\usepackage{enumitem}
\usepackage{array}
\usepackage{booktabs}
\usepackage{ragged2e}
\input{pkg_def}

\usepackage{flushend}

\usepackage[pagebackref=true,breaklinks=true,letterpaper=true,colorlinks,bookmarks=false]{hyperref}

\avssfinalcopy 

\def\avssPaperID{45} 

\ifavssfinal\fi
\begin{document}

\title{Real Image Super-Resolution using GAN through modeling of LR and HR process}

\author{Rao Muhammad Umer,\\
Institute of AI for Health (AIH),\\
Helmholtz Munich, Germany.\\
{\tt\small engr.raoumer943@gmail.com}
\and
Christian Micheloni,\\
Department of Mathematics and Computer Science,\\
University of Udine, Italy.\\
{\tt\small christian.micheloni@uniud.it}
}

\maketitle
\thispagestyle{empty}

\begin{abstract}
  The current existing deep image super-resolution methods usually assume that a Low Resolution (LR) image is bicubicly downscaled of a High Resolution (HR) image. However, such an ideal bicubic downsampling process is different from the real LR degradations, which usually come from complicated combinations of different degradation processes, such as camera blur, sensor noise, sharpening artifacts, JPEG compression, and further image editing, and several times image transmission over the internet and unpredictable noises. It leads to the highly ill-posed nature of the inverse upscaling problem. To address these issues, we propose a GAN-based SR approach with learnable adaptive sinusoidal nonlinearities incorporated in LR and SR models by directly learn degradation distributions and then synthesize paired LR/HR training data to train the generalized SR model to real image degradations. We demonstrate the effectiveness of our proposed approach in quantitative and qualitative experiments. 
\end{abstract}
\let\thefootnote\relax\footnote{978-1-6654-6382-9/22/\$31.00 ©2022 IEEE}
\section{Introduction}
Single image super-resolution (SISR) aims to restore the high-resolution (HR) image from its low-resolution (LR) image counterpart. SISR problem is a fundamental low-level vision and image processing problem with various practical applications in \eg, satellite imaging, medical imaging, astronomy, remote sensing, surveillance, image compression, environment and climate change monitoring, mobile photography, image / video enhancement, and security and surveillance imaging, etc. With the increasing amount of HR images / videos data on the internet, there is a great demand for storing, transferring, and sharing such large sized data with low cost of storage and bandwidth resources. Moreover, the HR images are usually downscaled to easily fit into display screens with different resolution while retaining visually plausible information. The downscaled LR counterpart of the HR can efficiently utilize lower bandwidth, storage save, and easily fit to various digital displays. However, some details are lost and sometimes visible artifacts appear when users downscale and upscale the digital contents.
\begin{figure}[t!]
\centering
\includegraphics[scale=0.9]{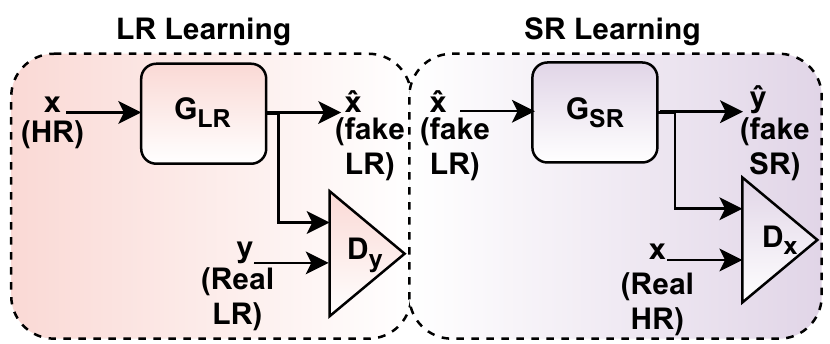}
\caption{The structure of our proposed real-world SR approach setup. In the LR Learning part, we train the LR generator network $\mathbf{G}_{LR}$ in a GAN framework, where our goal is to learn the real LR ($\by$) corruptions/degradations. Then, we use the synthesized paired LR/HR data by the $\mathbf{G}_{LR}$ model to train the generalized SR model $\mathbf{G}_{SR}$ in the SR Learning part. Both the $\mathbf{G}_{LR}$ and $\mathbf{G}_{SR}$ generators utilize the modified residual structure (refer to the sections~\ref{sec:lr_learning} and \ref{sec:sr_learning} for more details).}
\label{fig:srrescsingan}
\vspace{-0.5cm}
\end{figure}

Mathematically, SISR is described as a linear forward observation model~\cite{umer2020srrescgan,umer2020srrescycgan} with the following image degradation process:
\begin{equation}
    \by = (\bH \otimes \Tilde{\bx})\downarrow_s + \eta,
    \label{eq:degradation_model}
\end{equation}
where, $\by$ is an observed LR image, $\bH$ is a \emph{down-sampling operator} (unknown) that convolves ($\otimes$) with a latent HR image $\Tilde{\bx}$ and resizes it by a scaling factor $s$, and $\eta$ is considered as an i.i.d additive white Gaussian noise (AWGN) of variance $\sigma^2$, \ie,~$\eta \sim \mathcal{N}\left(0, \sigma^{2}\right)$. However, in real-world settings,  $\eta$ also accounts for all possible errors during the image acquisition process that include inherent sensor noise, stochastic noise, compression artifacts, and the possible mismatch between the forward observation model and the camera device. The operator $\bH$ is usually ill-conditioned or singular due to the presence of unknown noise realization ($\eta$) that turn the SISR to a highly ill-posed nature of inverse problems. Since, due to ill-posed nature, there are many possible solutions thus regularization is required to select the most plausible ones.

Recently, numerous works have been addressed towards the task of SISR~\cite{kim2016vdsrcvpr,Lim2017edsrcvprw,kai2017ircnncvpr,kai2018srmdcvpr,yuan2018unsupervised,Li2019srfbncvpr,zhang2019deep,srwdnet,Umer_2020_ICPR,luo2020unfoldingsr,zhou2020crossgraphsr,li2021laparsr,gou2020clearer} and real-world SISR~\cite{ledig2017srgan,wang2018esrgan,lugmayr2019unsupervised,fritsche2019dsgan,umer2020srrescgan,umer2020srrescycgan,Umer2109srresstargan}. Most of the SISR methods assume usually bicubic downsampling process, which is different from the real LR degradations. The real-world SISR methods try to solve the problem by utilizing data distribution learning using the GAN~\cite{goodfellow2014gan} framework. However, they do not generalize well to the real complex degradation, which usually come from the complicated degradation processes, \ie, sensor noise, camera blur, sharping artifacts, JPEG compression, and further image editing, and several times image transmission over the internet. In the most recent works~\cite{wang2021realesrgan,zhang2021designingpd}, the authors aim to restore general real-world LR images by synthesizing training pairs with a more practical degradation process. As the real-world degradation space is much larger/complex,
the synthetic modeling also becomes challenging. Moreover, the generators (\ie, LR/HR) require a more powerful capability to model the complex training data, while the gradients needs to be more accurate for local detail enhancement with some sophisticated nonlinearities inside the network. 

In this work, we proposed the GAN-based real image SR approach that solves the problem by modeling the LR/HR process with adaptive sinusoidal activitions (\ie,~better represent the complicated signals) and thus synthesize the more realistic paired LR/HR data to train the generalized SR model for the real SR task. The structure of our proposed real-world SR approach setup is shown in Fig.~\ref{fig:srrescsingan}. In the LR learning, we train the LR network ($\mathbf{G}_{LR}$) with modified residual structure (\ie, incorporating the sinusodial non-linearities) in a GAN-framework~\cite{goodfellow2014gan} to generate the realistic LR images as the corruptions/degradations of the real LR images ($\by$). After that, we use the synthesized paired LR/HR data to train the generalized SR model in the SR Learning part. The SR network ($\mathbf{G}_{SR}$) is trained in a GAN-framework~\cite{goodfellow2014gan} with the modified residual structure to super-resolve the LR images.   

We evaluate our proposed SR method on the Real-World Super-resolution (RWSR) dataset~\cite{NTIRE2020RWSRchallenge} to show the effectiveness of our approach through the quantitative and qualitative experiments. We summarize our contributions in three fold as:
\begin{enumerate}
    \item We propose an end-to-end deep SRResCSinGAN for the real-world SR task. Instead of using traditional bicubic downsampling or the existing deep LR degradation methods, we synthesize the paired training data with a more practical image corruptions/degradations by modeling the LR/HR process.
    \item By exploiting the sinusoidal non-linearities, we employ the modified residual network structure incorporated in both LR and SR learning stages, which better models the underlying complex signals \ie,~real LR and HR process.
    \item Our proposed approach achieve better quantitative and visual performance in terms of PSNR/SSIM/LPIPS (refer to Tables~\ref{tab:comp_sota_srrescsingan} and ~\ref{tab:ablation_table}).
\end{enumerate}

\begin{figure*}[!t]
\centering
\includegraphics[scale=0.70]{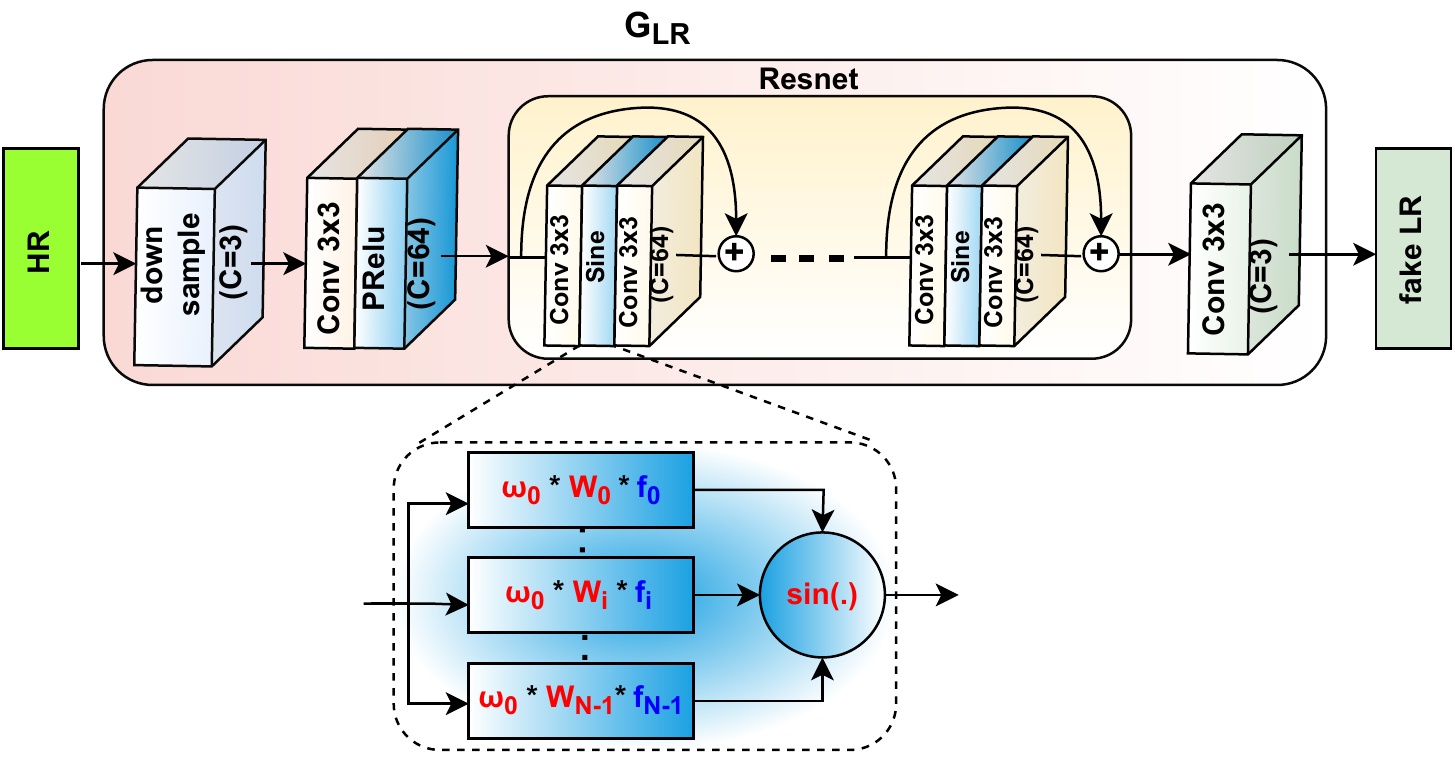}
\caption{The generator architecture of proposed LR learning stage. The $C$ denotes the output feature channels. Inside the \emph{Sine} layer, ${\color{red}\omega_0}$ (hyperparameter) is the scalar frequency factor, ${\color{red}\mathbf{W}}$ are the learnable sine weights, and ${\color{blue}f}$ are the \emph{Conv} layer feature maps (refer to section~\ref{sec:resnet_sine} for more details of \emph{Sine} layer).}
\label{fig:lr_learning_srrescsingan}
\vspace{-0.3cm}
\end{figure*}

\section{Related Work}
\subsection{Real World SISR methods}
Recently, numerous works~\cite{kim2016vdsrcvpr,Lim2017edsrcvprw,kai2017ircnncvpr,kai2018srmdcvpr,yuan2018unsupervised,Li2019srfbncvpr,zhang2019deep,srwdnet,Umer_2020_ICPR,luo2020unfoldingsr,zhou2020crossgraphsr,li2021laparsr,gou2020clearer} have addressed the task of SISR using deep CNNs for their powerful feature representation capabilities. The SISR methods mostly rely on the PSNR-based metric by optimizing the $\mathcal{L}_1$/$\mathcal{L}_2$ losses with blurry results in a supervised way, while they do not preserve the visual quality with respect to human perception. Moreover, the above-mentioned methods are deeper or wider CNN networks to learn non-linear mapping from LR to HR with the ideal bicubic downsampling, while neglecting the real-world settings. 

For the real image SR task, several attempts~\cite{ledig2017srgan,wang2018esrgan,lugmayr2019unsupervised,fritsche2019dsgan,umer2020srrescgan,umer2020srrescycgan,Umer2109srresstargan} have done to solve for realistic LR degradation. However, the real SR methods still suffer unpleasant artifacts and challenging for learning fine-grained corruptions/degradations with unpaired data. Our approach takes into account the real-world settings by increasing its applicability in practical scenarios. 

\subsection{Blind / Non-Blind degradation models}
Classical degradation model (refer to Eq.~\eqref{eq:degradation_model}) is mostly used in the blind / non-blind deep SISR methods. The common choice, in the existing SISR degradation models, usually consist of a sequence of blur kernel (\ie, Gaussian/motion), downsampling (\ie, bicubic, bilinear, nearest-neighbor), and
noise addition (\ie, AWGN). In the existing deep SISR methods~\cite{wang2021realesrgan,zhang2021designingpd}, they attempt to explicit model the real-world degradation to super-resolve the real LR images. But, yet the real-world degradations are too complex to be explicitly modeled. Therefore, implicit modeling using GAN framework within the network is a suitable choice to synthesize more practical degradations.

\section{Proposed Method}
\subsection{Problem Formulation}
By referencing to the Eq.~\eqref{eq:degradation_model}, the recovery of $\bx$ from $\by$ mostly relies on the variational approach for combining the observation and prior knowledge, and is given by the following objective function:
\begin{equation}
    \bx^* = \underset{\mathbf{x}}{\arg \min }~\frac{1}{2\sigma^2}\|\by - (\bH \otimes \bx)\downarrow_s\|_2^{2}+\lambda \phi(\bx),
    \label{eq:eq1}
\end{equation}
where, $\frac{1}{2\sigma^2}\|\by - (\mathbf{H}\otimes\bx)\downarrow_s\|_2^2$ is the data fidelity (also known as log-likelihood) term that measures the closeness of the solution to the observations, $\phi(\bx)$ is the regularization term that encodes the image prior knowledge, and $\lambda$ is the trade-off parameter that governs the compromise between the data fidelity and the regularizer term. Interestingly, the variational approach has a direct link to the Bayesian approach and the derived solutions can be described either as penalized maximum likelihood or as maximum a posteriori (MAP) estimates~\cite{bertero1998map1,figueiredo2007map2}. Thanks to the recent advances of deep learning, the regularizer (\ie,~$\phi(\bx)$) is employed by the SRResCGAN~\cite{umer2020srrescgan} generator structure that is inspired by a powerful image regularization and large-scale optimization techniques to solve the real-world SISR task.

\subsection{Residual Network (Resnet) with adaptive Sinusoidal non-linearities}
\label{sec:resnet_sine}
Over the past decades, numerous works have investigated a variety of possible activation functions, such as sigmoid, ReLU, Tanh, PReLU, RBF, and many more to model the natural images. The preferred choice that has emerged over the years is the ReLU activation unit due to promoting sparsity of the feature maps and the faster training of very deep networks. The continuous and piecewise linear functions have proven as a universal approximation of complex signals such as natural images. Recent works have demonstrated the potential to robustly outperform ReLU and other non-linearities by using alternative activation functions for image reconstruction / restoration tasks, such as deep spline activations~\cite{unser2019splines} and periodic nonlinearities like sinusoidal~\cite{sitzmann2019siren}. Motivated by the continuous and differentiable periodic nonlinearities (\ie,~sinusoidal) that are capable of representing complex and fine details of signals better than the ReLU and others, we exploit the sinusoidal nonlinearities incorporated in the modified structure of deep residual network (\emph{Resnet}).

We describe the overall explicit compositional structure of the L-layer deep residual network (\emph{Resnet}) with the following formulation:
\begin{equation}
\label{eq:DeepNet}
    \begin{split}
        \mathbf{f}_{resnet}(\bx) = \left(\left({\color{blue}f_L} \circ {\color{red}\sigma_L} \circ {\color{blue}f_{L-1}}\right)(\bx_{L-1}) + \bx_{L-1}\right) \circ \cdots \circ \\
        \left(\left({\color{blue}f_2} \circ {\color{red}\sigma_1} \circ {\color{blue}f_1}\right)(\bx) + \bx \right),
    \end{split}
\end{equation}
Here, ${\color{blue}f}$ is the affine transformation (\ie,~\emph{Conv} layer) defined by the weight matrix ${\color{blue}\mathbf{W}}$ and the biases ${\color{blue}\mathbf{b}}$ applied to the input as: 
\begin{equation}
{\color{blue}f}(\bx) =  {\color{blue}\mathbf{W}} * \mathbf{x} + {\color{blue}\mathbf{b}}
\end{equation}
And, followed by the sine nonlinearity~\cite{sitzmann2019siren} ${\color{red}\sigma}$ applied to the resulting vector ${\color{blue}f}$ as:
\begin{equation}
{\color{red}\sigma}({\color{blue}f}) = {\color{red}sin} \left({\color{red}\omega_0} . {\color{red}\mathbf{W}} {\color{blue}f} \right)
\end{equation}
where, ${\color{red}\omega_0}$ is the scalar frequency factor, which is a hyperparameter. The derivative of the sine is a cosine (\ie, the phase-shifted sine) for the backpropagation. The weights of the \emph{Sine} layer are updated during the training via the stochastic gradient descent steps by minimization of the loss function. To initialize the weights (${\color{red}\mathbf{W}}$) of the \emph{Sine} layer, we use the same initialization technique as done in \cite{sitzmann2019siren}, where we draw the weights with ${\color{red}\mathbf{W}_i} \sim \mathcal{U}(-\sqrt{6/n}, \sqrt{6/n})$ which ensures that the input to each sine activation is normal distributed with a unit standard deviation. 
\begin{figure*}[h!]
\centering
\includegraphics[scale=0.55]{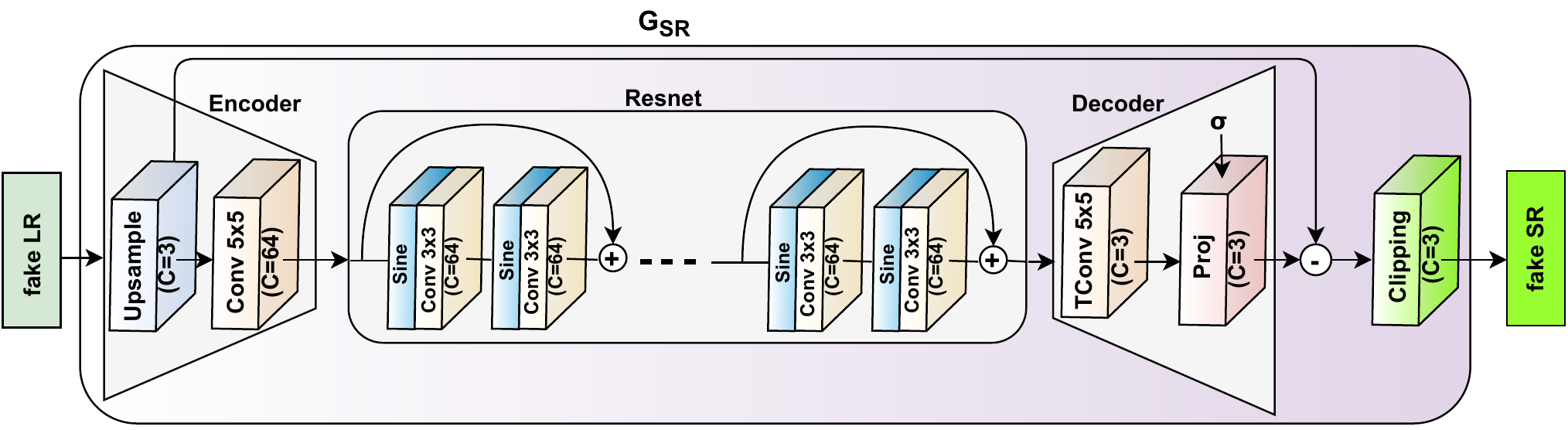}
\caption{The generator architecture of proposed SR learning stage. The $C$ denotes the output feature channels. The \emph{Sine} layer denotes sinusoidal nonlinearities (refer to Fig.~\ref{fig:lr_learning_srrescsingan} and section~\ref{sec:resnet_sine} for more details). The $\sigma$ is the trainable projection layer parameter.}
\label{fig:sr_learning_srrescsingan}
\vspace{-0.5cm}
\end{figure*}
\section{LR Learning Model}
\label{sec:lr_learning}
In the LR learning phase, we train the model ($\mathbf{G}_{LR}$) in a GAN framework as shown in Fig.~\ref{fig:lr_learning_srrescsingan}. In the next sections \ref{sec:lr_gen_arch}, \ref{sec:lr_gen_loss}, and \ref{sec:lr_train_desc}, we describe the network architectures, training losses, and other training details. 

\subsection{Network Architectures}
\label{sec:lr_gen_arch}
The modified LR generator network ($\mathbf{G}_{LR}$)  (as shown in Fig.~\ref{fig:lr_learning_srrescsingan}) consists of 8 \emph{Resnet} blocks. Each residual block contains the \emph{Sine} (\ie,~sinusoidal nonlinearities) layer that is sandwiched between two \emph{Conv} layers. All \emph{Conv} layers have $3\times3$ kernel support with 64 feature maps. Finally, \emph{sigmoid} nonlinearity is applied on the output of the $\mathbf{G}_{LR}$ network.  While, the LR discriminator network ($\mathbf{D}_{\by}$) consists of a three-layer convolutional network that operates on a patch level~\cite{isola2017image, li2016precomputed}. All \emph{Conv} layers have $5\times5$ kernel support with feature maps from 64 to 256 and also applied Batch Normalization and Leaky ReLU (LReLU) activations after each \emph{Conv} layer except the last \emph{Conv} layer that maps 256 to 1 features. It is trained to discriminate the real LR images ($\by$) from the fake LR images ($\hat{\bx}$) generated by the $\mathbf{G}_{LR}$.

\subsection{Network Losses}
\label{sec:lr_gen_loss}
To learn the degradation/corruptions from the LR domain ($\by$) images, we train the modified network $\mathbf{G}_{LR}$ in a GAN framework~\cite{goodfellow2014gan} with the following loss functions:
\begin{equation}
    \mathcal{L}_{\mathbf{G}_{LR}} = \mathcal{L}_{color} + 0.005 \cdot \mathcal{L}_{tex} + 0.01 \cdot \mathcal{L}_{per}
    \label{eq:dsgan_lg}
\end{equation}
where, these loss functions are defined as follows:\\
\textbf{Color loss ($\mathcal{L}_{color}$):} It is basically the $\mathcal{L}_1$ loss which focuses on the low frequencies of the image.
\begin{equation}
    \mathcal{L}_{color} = \frac{1}{N} \sum_{i=1}^{N}\left\|\mathbf{w}_{\mathrm{L}} * \mathbf{G}_{LR}(\mathbf{x}^{(i)})-\mathbf{w}_{\mathrm{L}} * \mathbf{x}^{(i)}\downarrow_s\right\|_{1}
\end{equation}
Here, $\mathbf{w}_{\mathrm{L}}$ is the low-pass filter, $N$ is the mini-batch size, and $\downarrow_s$ is the downscaling factor.\\
\textbf{Texture loss ($\mathcal{L}_{tex}$):} It focuses on the high frequencies of the image.
\begin{equation}
    \mathcal{L}_{tex} = \frac{1}{N} \sum_{i=1}^{N} \operatorname{mean}\left(\log \mathbf{D}_{\by}\left(\mathbf{w}_{\mathrm{H}} *\mathbf{G}_{LR}(\mathbf{x}^{(i)}) \right)\right) 
\end{equation}
Here, $\mathbf{w}_{\mathrm{H}}$ is the high-pass filter.\\
\textbf{Perceptual loss ($\mathcal{L}_{\mathrm{per}}$):} It focuses on the perceptual quality of the output image and is defined as:
\begin{equation}
    \mathcal{L}_{\mathrm{per}}=\frac{1}{N} \sum_{i=1}^{N}\mathcal{L}_{\mathrm{VGG}}=\frac{1}{N} \sum_{i=1}^{N}\|\phi(\mathbf{G}_{LR}(\bx^{(i)}))-\phi(\bx^{(i)}\downarrow_s)\|_{1}
\end{equation}
where, $\phi$ is the feature extracted from the pretrained VGG network as done in DSGAN~\cite{fritsche2019dsgan}.\\

\subsection{Training description}
\label{sec:lr_train_desc}
We train the $\mathbf{G}_{LR}$ network with image patches $512\times512$, which are bicubically downsampled with MATLAB \emph{imresize} function. We randomly crop the LR domain images ($\by$) by $128\times128$ as done in ~\cite{fritsche2019dsgan}. We set the $\omega_0=30$ for the \emph{Sine} layer. We train the network for 300 epochs with a batch size of 16 using Adam optimizer~\cite{Kingma2015AdamAM} with parameters $\beta_1 =0.5$, $\beta_2=0.999$, and $\epsilon=10^{-8}$ without weight decay for both generator and discriminator to minimize the loss in \eqref{eq:dsgan_lg}. The learning rate is initially set to $2.10^{-4}$ for the first 150 epochs and then linearly decayed to zero after the remaining (\ie, 150) epochs as done in ~\cite{fritsche2019dsgan}.

\section{SR Learning Model}
\label{sec:sr_learning}
In the SR learning phase, we train the model ($\mathbf{G}_{SR}$) in a GAN framework as shown in Fig.~\ref{fig:sr_learning_srrescsingan}. In the next sections \ref{sec:sr_gen_arch}, \ref{sec:sr_gen_loss}, and \ref{sec:sr_train_desc}, we describe the network architectures, training losses, and other training details. 

\subsection{Network Architectures}
\label{sec:sr_gen_arch}
We use the SR generator $\mathbf{G}_{SR}$  network (as shown in Fig.~\ref{fig:sr_learning_srrescsingan}) which is an \emph{Encoder-Resnet-Decoder} like structure as done in SRResCGAN~\cite{umer2020srrescgan} with the modified \emph{Resnet} structure by incorporating the sinusoidal nonlinearities. In the $\mathbf{G_{SR}}$ network, both \emph{Encoder} and \emph{Decoder} layers have $64$ convolutional feature maps of $5\times5$ kernel size with $C \times H\times W$ tensors, where $C$ is the number of channels of the input image. Inside the \emph{Encoder}, LR image is upsampled by the Bicubic kernel with \emph{Upsample} layer. \emph{Resnet} consists of $5$ residual blocks with two Pre-activation \emph{Conv} layers, each of $64$ feature maps with kernel support $3\times3$, and the preactivition is the \emph{Sine} layer with $64$ output feature channels. The trainable projection layer~\cite{Lefkimmiatis2018UDNet} inside the \emph{Decoder} computes the proximal map with the estimated noise standard deviation $\sigma$ and handles the data fidelity and prior terms. The noise realization is estimated in the intermediate \emph{Resnet} that is sandwiched between \emph{Encoder} and \emph{Decoder}. The estimated residual image after \emph{Decoder} is subtracted from the LR input image. Finally, the clipping layer incorporates our prior knowledge about the valid range of image intensities and enforces the pixel values of the reconstructed image to lie in the range $[0, 255]$. The reflection padding is also used before all \emph{Conv} layers to ensure slowly varying changes at the boundaries of the input images.

The SR discriminator network ($\mathbf{D}_{\bx}$) is trained to discriminate the real HR images ($\bx$) from the fake HR images ($\hat{\by}$) generated by the $\mathbf{G}_{SR}$. The raw discriminator network contains 10 convolutional layers with kernels that support $3\times3$ and $4\times4$ of increasing feature maps from $64$ to $512$ followed by Batch Normalization and leaky ReLU as done in SRGAN~\cite{ledig2017srgan}.

\subsection{Network Losses}
\label{sec:sr_gen_loss}
To learn the image super-resolution for the HR domain ($\bx$) images, we train the modified network $\mathbf{G}_{SR}$ in a GAN framework with the following loss functions:
\begin{equation}
    \mathcal{L}_{G_{SR}} = \mathcal{L}_{\mathrm{per}}+ \mathcal{L}_{\mathrm{GAN}} + \mathcal{L}_{tv} + 10\cdot \mathcal{L}_{\mathrm{1}}
    \label{eq:l_g}
\end{equation}
where, these loss functions are defined as follows:\\
\textbf{Perceptual loss ($\mathcal{L}_{\mathrm{per}}$):} It focuses on the perceptual quality of the output image and is defined as:
\begin{equation}
    \mathcal{L}_{\mathrm{per}}=\frac{1}{N} \sum_{i=1}^{N}\mathcal{L}_{\mathrm{VGG}}=\frac{1}{N} \sum_{i=1}^{N}\|\phi(\mathbf{G}_{SR}(\hat{\bx}^{(i)}))-\phi(\bx^{(i)})\|_{1}
\end{equation}
where, $\phi$ is the feature extracted from the pretrained VGG-19 network as done in ESRGAN~\cite{wang2018esrgan}.\\
\textbf{GAN loss ($\mathcal{L}_{\mathrm{GAN}}$):} It focuses on the high frequencies of the output image and it is defined as:
\begin{equation}
   \mathbf{D}_{\bx}(\bx, \hat{\by})(C) = \sigma(C(\bx)-\mathbb{E}[C(\hat{\by})])
\end{equation}
Here, $C$ is the raw discriminator output and $\sigma$ is the sigmoid function. By using the relativistic discriminator~\cite{wang2018esrgan}, we have:
\begin{equation}
    \begin{split}
       \mathcal{L}_{\mathrm{GAN}} = \mathcal{L}_{\mathrm{RaGAN}} = &-\mathbb{E}_{\bx}\left[\log \left(1-\mathbf{D}_{\bx}(\bx, \mathbf{G}_{SR}(\hat{\bx}))\right)\right] \\
    &-\mathbb{E}_{\hat{\by}}\left[\log \left(\mathbf{D}_{\bx}(\mathbf{G}_{SR}(\hat{\bx}), \bx)\right)\right] 
    \end{split}
\end{equation}
where, $\mathbb{E}_{\bx}$ and $\mathbb{E}_{\hat{\by}}$ represent the operations of taking average for all real HR ($\bx$) images and fake HR ($\hat{\by}$) images in the mini-batches, respectively.\\ 
\textbf{TV (total-variation) loss ($\mathcal{L}_{tv}$):} It focuses to minimize the gradient discrepancy and produces sharpness in the output SR image, and it is defined as:
\begin{equation}
\begin{split}
   \mathcal{L}_{tv}=&\frac{1}{N} \sum_{i=1}^{N}\left(\|\nabla_{h} \mathbf{G}_{SR}(\hat{\bx}^{(i)}) - \nabla_{h} (\bx^{(i)})\|_{1}+ \right. \\ & \left \|\nabla_{v} \mathbf{G}_{SR}(\hat{\bx}^{(i)}) - \nabla_{v} (\bx^{(i)}) \|_{1}\right).
\end{split}
\end{equation}
Here, $\nabla_{h}$ and $\nabla_{v}$ denote the operators calculating the image directional derivatives in the horizontal and vertical directions, respectively. \\
\textbf{Content loss ($\mathcal{L}_{\mathrm{1}}$):} It is defined as:
\begin{equation}
    \mathcal{L}_{1} = \frac{1}{N} \sum_{i=1}^{N} \|\mathbf{G}_{SR}(\hat{\bx}^{(i)}) - \bx^{(i)}\|_{1}
\end{equation}
where, $N$ represents the size of mini-batch.

\subsection{Training description}
\label{sec:sr_train_desc}
During the training phase, we set the input LR patch size as $32\times32$. We train the network for 51000 training iterations with a batch size of 16 using Adam optimizer with parameters $\beta_1 =0.9$, $\beta_2=0.999$, and $\epsilon=10^{-8}$ without weight decay for both generator and discriminator to minimize the loss in \eqref{eq:l_g}.  We set the $\omega_0=30$ for the \emph{Sine} layer. The learning rate is initially set to $10^{-4}$ and then multiplies by $0.5$ after 5K, 10K, 20K, and 30K iterations. The projection layer parameter $\sigma$ is estimated according to \cite{liu2013single} from the input LR image. We initialize the projection layer parameter $\alpha$ on a log-scale values from $\alpha_{max}=2$ to $\alpha_{min}=1$ and then further fine-tune during the training via back-propagation.

\begin{table*}[t]
	\centering
	\tabcolsep=0.01\linewidth
	\scriptsize
	\resizebox{0.65\textwidth}{!}{
	\begin{tabular}{l|l|c|c|c|c}
	    \toprule
		\textbf{Dataset (LR/HR pairs)} & \textbf{SR methods} & \textbf{\#Params} & \textbf{PSNR$\uparrow$} & \textbf{SSIM$\uparrow$} & \textbf{LPIPS$\downarrow$} \\ 
		\midrule
		 &  &  & \multicolumn{3}{c}{sensor noise ($\sigma=8$)} \\
		 \cline{4-6}
		Bicubic & EDSR~\cite{Lim2017edsrcvprw} & $43M$ & 24.48 & 0.53 & 0.6800 \\
		Bicubic & ESRGAN~\cite{wang2018esrgan} & $16.7M$ & 17.39 & 0.19 & 0.9400 \\
		CycleGAN~\cite{lugmayr2019unsupervised} & ESRGAN-FT~\cite{lugmayr2019unsupervised} & $16.7M$ & 22.42 & 0.55 & 0.3645 \\
		DSGAN~\cite{fritsche2019dsgan} &  ESRGAN-FS~\cite{fritsche2019dsgan} & $16.7M$ & 22.52 & 0.52 & {\color{red}0.3300} \\
		DSGAN~\cite{fritsche2019dsgan} & SRResCGAN~\cite{umer2020srrescgan} & $380K$ & {\color{blue}25.46} & {\color{blue}0.67} & {\color{blue}0.3604} \\
		DSSinGAN (ours) & SRResCSinGAN (ours) & $380K$ & {\color{red}25.50} & {\color{red}0.69} & 0.3750 \\
		\midrule
		&  &  & \multicolumn{3}{c}{JPEG compression (quality=30)} \\
		\cline{4-6}
		Bicubic & EDSR~\cite{Lim2017edsrcvprw} & $43M$ & {\color{red}23.75} & {\color{blue}0.62} & 0.5400 \\
		Bicubic & ESRGAN~\cite{wang2018esrgan} & $16.7M$ & 22.43 & 0.58 & 0.5300 \\
		CycleGAN~\cite{lugmayr2019unsupervised} & ESRGAN-FT~\cite{lugmayr2019unsupervised} & $16.7M$ & 22.80 & 0.57 & {\color{red}0.3729} \\
		DSGAN~\cite{fritsche2019dsgan} & ESRGAN-FS~\cite{fritsche2019dsgan} & $16.7M$ & 20.39 & 0.50 & {\color{blue}0.4200} \\
		DSGAN~\cite{fritsche2019dsgan} & SRResCGAN~\cite{umer2020srrescgan}  & $380K$ & 23.34 & 0.59 & 0.4431 \\
		DSSinGAN (ours) & SRResCSinGAN (ours) & $380K$ & {\color{blue}23.70} & {\color{red}0.63} & 0.4258  \\
		\midrule
		&  &  & \multicolumn{3}{c}{unknown (validset)~\cite{NTIRE2020RWSRchallenge}} \\
		\cline{4-6}
		DSGAN~\cite{fritsche2019dsgan} & SRResCGAN~\cite{umer2020srrescgan}  & $380K$ & 25.05 & 0.67 & {\color{red}0.3357} \\
		DSSinGAN (ours) & SRResCSinGAN (ours) & $380K$ & {\color{blue}25.58} & {\color{blue}0.69} & {\color{blue}0.3610} \\
		DSSinGAN (ours) & SRResCSinGAN+ (ours) & $380K$ & {\color{red}25.89} & {\color{red}0.71} & 0.3769 \\
		\bottomrule
	\end{tabular}}
	\vspace{-0.3cm}
	\caption{$\times4$ SR quantitative results comparison of our method over the DIV2K validation-set (100 images) with added two known degradations~\ie,~sensor noise ($\sigma=8$) and JPEG compression ($quality=30$) artifacts. Bottom section: $\times4$ SR results comparison with the unknown corruptions in the RWSR challenge series (validation-set)~\cite{NTIRE2020RWSRchallenge}. The arrows indicate if high $\uparrow$ or low $\downarrow$ values are desired. The best performance is shown in {\color{red} red} and the second best performance is shown in {\color{blue} blue}.}
	\label{tab:comp_sota_srrescsingan}
	\vspace{-0.5cm}
\end{table*}

\section{Experimental Results}
\subsection{Training data preparation}
\label{sec:train_data}
We use the LR domain images ($\by$: 2650 HR images)  that are corrupted with unknown degradation, e.g., sensor noise, compression artifacts, sharpening artifacts, etc., and HR domain images ($\bx$: 800 clean HR images), provided in the NTIRE-2020 Real-World Super-resolution (RWSR)  Challenge~\cite{NTIRE2020RWSRchallenge}. The LR domain images contain synthetic visible corruptions that are similar to the induced corruptions by the current camera devices. We use the LR and HR domain data to train the $\mathbf{G}_{LR}$ network to learn the domain degradation/corruptions. Then, we train the $\mathbf{G}_{SR}$ network by synthesizing the realistic LR/HR paired training data.

\subsection{Data augmentation}
We take the LR/HR patches due to the network training efficiency and we also assume that the patch based degradation is same as in the whole image. We augment the training data with random vertical and horizontal flipping, and $90^{\circ}$ rotations. Moreover, we also consider another effective data augmentation technique, called  mixture of augmentation (MOA) \cite{yoo2020rethinking} strategy. In the MOA, a data augmentation (DA) method, among \ie, Blend, RGB permutation, Mixup, Cutout, Cutmix, Cutmixup, and CutBlur is randomly selected then applied on the inputs. This MOA technique encourages the network to acquire more generalization power by partially blocking or corrupting the training sample. 

\subsection{Evaluation metrics}
We evaluate the trained model under the Peak Signal-to-Noise Ratio (PSNR), Structural Similarity (SSIM), and LPIPS metrics. The PSNR and SSIM are distortion-based measures that correlate poorly with actual perceived similarity, while LPIPS better correlates with human perception than the distortion-based/handcrafted measures. As LPIPS is based on the features of pretrained neural networks, so we use it for the quantitative evaluation with features of AlexNet. The quantitative SR results are evaluated on the $RGB$ color space. To further enhance the fidelity, we use a self-ensemble strategy~\cite{timofte2016seven} (denoted as SRResCSinGAN+) at the test time, where the LR inputs are flipped/rotated and the SR results are aligned and averaged for enhanced prediction.

\subsection{Comparison with state-of-the-art SR methods}
\label{sec:comp_sota}
We compare our method with other state-of-art SR methods including EDSR~\cite{Lim2017edsrcvprw}, ESRGAN~\cite{wang2018esrgan}, ESRGAN-FT~\cite{lugmayr2019unsupervised}, and ESRGAN-FS~\cite{fritsche2019dsgan} and SRResCGAN~\cite{umer2020srrescgan}, whose source codes are available online. The two degradation settings (\ie, sensor noise, JPEG compression) have been considered under the same experimental situations for all methods. We run all original source codes and trained models by the default parameters settings for comparison. 

The EDSR is trained without perceptual loss (only $\mathcal{L}_{\mathrm{1}}$) by a deep SR residual network using bicubic supervision. The ESRGAN is trained with the $\mathcal{L}_{\mathrm{perceptual}}$, $\mathcal{L}_{\mathrm{GAN}}$, and  $\mathcal{L}_{\mathrm{1}}$ by a deep SR network using bicubic supervision. The ESRGAN-FT and ESRGAN-FS apply the same SR architecture and perceptual losses as in the ESRGAN using the two known degradation supervisions. The SRResCGAN is trained with the similar losses combination as done in the ESRGAN using the two known degradation supervisions. We train the proposed SRResCSinGAN with the similar losses combination as done in the ESRGAN and SRResCGAN with the modified Resnet structure by the sine nonlinearities.

Table~\ref{tab:comp_sota_srrescsingan} shows the quantitative results comparison of our method over the DIV2K validation-set (100 images) with two known degradations (\ie, sensor noise, JPEG compression) as well as the unknown degradation in the RWSR challenge dataset~\cite{NTIRE2020RWSRchallenge}. In the case of sensor noise, our method has better PSNR/SSIM values compared to all existing SR methods, while we have comparable LPIPS value. Since these are the contradicted measures (PSNR/SSIM vs. LPIPS), our objective is to achieve a good PSNR/SSIM score, while getting the satisfactory LPIPS value. In the case of jpeg compression artifacts, our proposed method has better PSNR/SSIM values except the EDSR, which is slightly better PSNR, but low LPIPS value and it has a very deep network with $43M$ parameters, while our model has only $380K$ parameters. Finally, in the case of unknown corruptions, our method has better SR results in terms of PSNR and SSIM, while we have comparable LPIPS value with others.   

\begin{figure}[h!]
    \centering
    \begin{subfigure}[t]{0.45\textwidth}
        \centering
        \includegraphics[width=\linewidth]{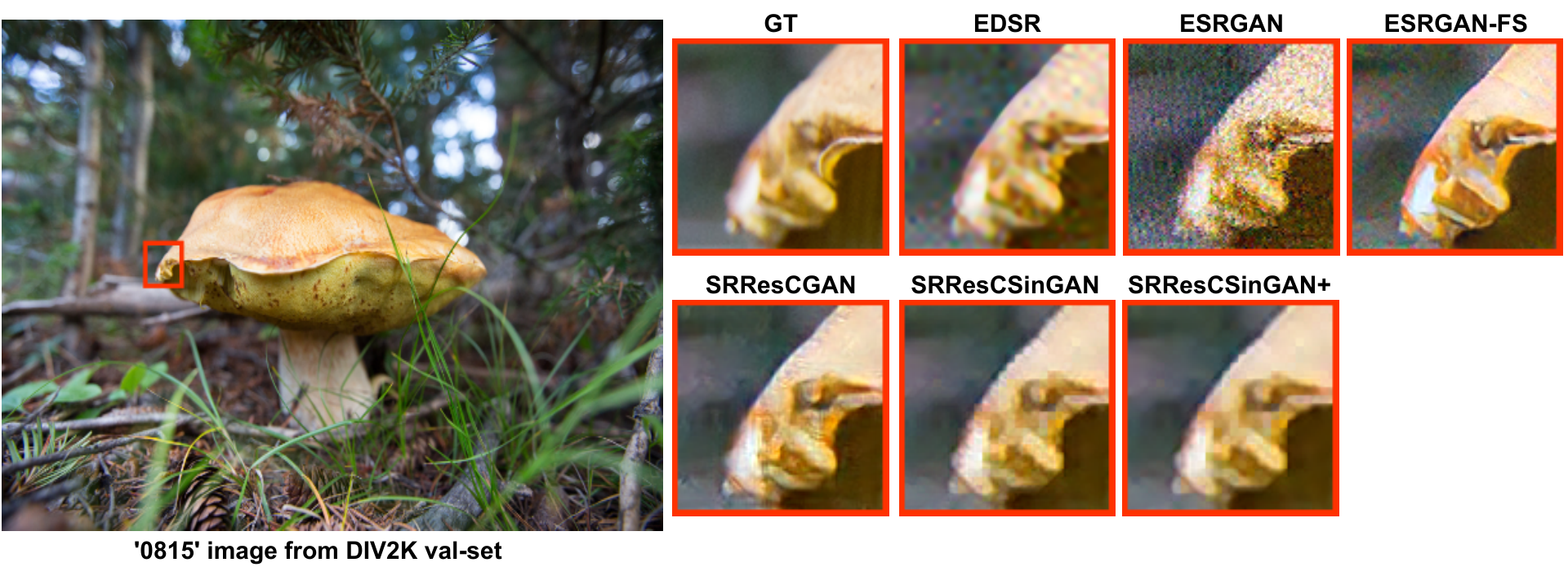}
    \end{subfigure}\\ 
    \begin{subfigure}[t]{0.45\textwidth}
        \centering
        \includegraphics[width=\linewidth]{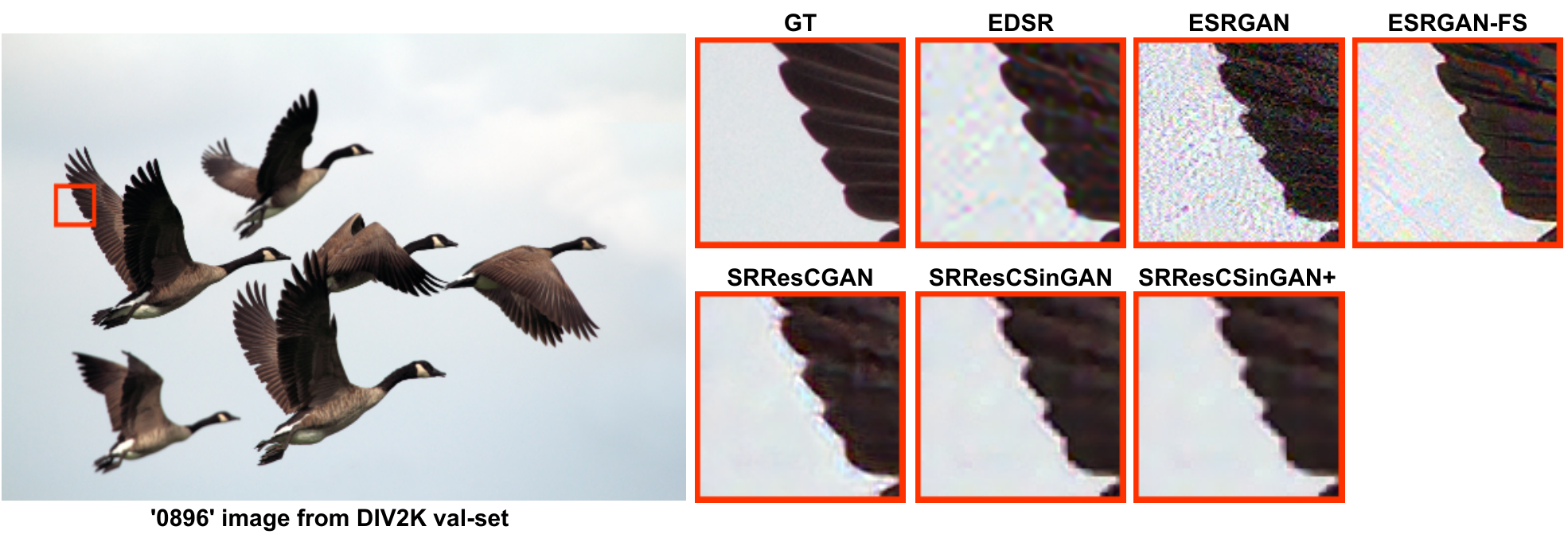}
    \end{subfigure}
    \caption{Visual SR comparison of our method with the other state-of-art methods on the DIV2K validation set at the $\times4$ upscaling factor.}
    \label{fig:4x_result_val}
    \vspace{-0.5cm}
\end{figure}

Regarding the visual quality, Fig.~\ref{fig:4x_result_val} shows the qualitative comparison of our method with the other SR methods on $\times 4$ upscaling factor (validation-set). In contrast to the existing state-of-art methods, our proposed method produces excellent SR results that are reflected in the PSNR/SSIM/LPIPS values, as well as the visual quality of the reconstructed images with almost no visible corruptions.

\begin{table}[h!]
	 \centering
	\tabcolsep=0.01\linewidth
	\scriptsize
	\resizebox{0.45\textwidth}{!}{
	\begin{tabular}{l|l|c|c|c}
		\toprule
		 \textbf{Dataset (LR/HR pairs)} & \textbf{SR method} & \textbf{PSNR$\uparrow$} & \textbf{SSIM$\uparrow$} & \textbf{LPIPS$\downarrow$} \\ 
		\midrule
		Bicubic & SRResCGAN & 24.13 & 0.57 & 0.4853  \\
		Bicubic & SRResCSinGAN & 24.78 & 0.62 & 0.4365  \\
		\midrule
		DSGAN & SRResCGAN & 25.05 & 0.67 & {\color{red}0.3357}  \\
		DSGAN & SRResCSinGAN & 25.53 &  {\color{blue}0.69} & 0.3792  \\
		\midrule
		DSSinGAN & SRResCSinGAN & {\color{blue}25.58} & {\color{blue}0.69} & {\color{blue}0.3610}  \\
		DSSinGAN & SRResCSinGAN+ & {\color{red}25.89} & {\color{red}0.71} & 0.3769  \\
		\bottomrule
	\end{tabular}}
	\vspace{-0.3cm}
	\caption{The quantitative SR results comparison of our method with others over the DIV2K validation set (100 images) with unknown degradation for our ablation study. The arrows indicate if high $\uparrow$ or low $\downarrow$ values are desired. The best performance is shown in {\color{red} red} and the second best performance is shown in {\color{blue} blue}.}
	\label{tab:ablation_table}
	\vspace{-0.5cm}
\end{table}
\subsection{Ablation Study}
For our ablation study, we generated different LR/HR pair data to train the SR models. We reached the better PSNR/SSIM score, while achieving good LPIPS for its better visual correlation with human perception. Table~\ref{tab:ablation_table} shows the quantitative results of our method over the DIV2K validation-set (100 images) with unknown degradation~\cite{NTIRE2020RWSRchallenge}. 

In the top section of the table, we trained the  SRResCGAN~\cite{umer2020srrescgan} method with and without sine nonlinearities with the bicubic downsampled data (refer to section~\ref{sec:sr_learning} for the SR learning training). The SRResCGAN with sine non-linearities (\ie, SRResCSinGAN) has achieved better results in terms of PSNR, SSIM, and LPIPS. 

In the middle section, we generated the LR data from the DSGAN~\cite{fritsche2019dsgan} as done in SRResCGAN~\cite{umer2020srrescgan} and then trained the two variants of our SR model with the generated LR/HR pairs. The SRResCSinGAN has better SR results in terms of PSNR and SSIM, while satisfactory  LPIPS value compared to the SRResCGAN. 

In the bottom section, we generated the LR data from the DSGAN with sine nonlinearities (denoted as DSSinGAN, refer to section~\ref{sec:lr_learning} for the LR learning) and then finally train our proposed SRResCSinGAN method with the generated LR/HR pairs. The SRResCSinGAN has better PSNR and LPIPS values, while the same SSIM value. To further enhance the performance, we used the self-ensemble strategy~\cite{timofte2016seven} at the test time, denoted as SRResCSinGAN+. It suggests that better generation of the LR images instead of the traditional bicubic downscaling gives the better performance gain and also incooperating the sinusoidal non-linearites instead of ReLU/PReLU activation in the resnet structure gives the improvement in the reconstruction quality.   

\section{Conclusion}
We proposed a deep SRResCSinGAN method for real image super-resolution by following the real-world settings. The proposed method solves the real image SR problem by implicitly modeling the degradation process within the network. The proposed approach first  synthesize the realistic paired training data with a more practical corruptions/degradations, instead of using the traditional bicubic downsampling or the existing deep learning based methods. Secondly, the proposed approach use the synthesized LR/HR paired data to train the generalized SR model to super-resolve the real LR images. The proposed approach incorporate the sinusoidal nonlinearities in the LR and HR model process to better representing the underlying complex signals in natural images. Our method achieves better SR results in terms of PSNR/SSIM values and comparable LPIPS values as well as visual quality compared to the existing state-of-art methods.

{\small
\bibliographystyle{ieee}
\bibliography{references}
}

\end{document}

%% file: pkg_def.tex
\def\bx{{\bf x}}
\def\by{{\bf y}}

\def\0{{\bf 0}}
\def\1{{\bf 1}}

\def\bH{{\bf H}}

\def\ie{\emph{i.e.}}
\def\eg{\emph{e.g.}}

